\documentclass{article}

\usepackage{arxiv}

\usepackage[utf8]{inputenc} 
\usepackage[T1]{fontenc}    
\usepackage{hyperref}       
\usepackage{url}            
\usepackage{booktabs}       
\usepackage{amsfonts}       
\usepackage{nicefrac}       
\usepackage{microtype}      
\usepackage{lipsum}
\usepackage{graphicx}
\graphicspath{ {./images/} }

\title{Fine-Tuning LLMs on Small Medical Datasets: Text Classification and Normalization Effectiveness on Cardiology reports and Discharge records}

\author{
 Noah Losch \\
  \\
  Institute of Medical Informatics \\
  University of Münster, Münster, Germany \\
  \texttt{noah.losch@uni-muenster.de} \\
   \And
 Lucas Plagwitz \\
  \\
  Institute of Medical Informatics \\
  University of Münster, Münster, Germany \\
  \texttt{lucas.plagwitz@uni-muenster.de} \\
   \And
 Antonius Büscher \\
  \\
  Institute of Medical Informatics\\ Department for Cardiology II-Electrophysiology \\
  University of Münster, Münster, Germany \\
  \texttt{antonius.buescher@uni-muenster.de} \\
   \And
 Julian Varghese \\
  \\
  Institute of Medical Informatics \\
  University of Münster, Münster, Germany \\
  \texttt{julian.varghese@uni-muenster.de}
}

\begin{document}
\maketitle
\begin{abstract}
We investigate the effectiveness of fine-tuning large language models (LLMs) on small medical datasets for text classification and named entity recognition tasks. Using a German cardiology report dataset and the i2b2 Smoking Challenge dataset, we demonstrate that fine-tuning small LLMs locally on limited training data can improve performance achieving comparable results to larger models. Our experiments show that fine-tuning improves performance on both tasks, with notable gains observed with as few as 200-300 training examples. Overall, the study highlights the potential of task-specific fine-tuning of LLMs for automating clinical workflows and efficiently extracting structured data from unstructured medical text.
\end{abstract}


\section{Introduction}
Large Languages Models (LLMs) have found many applications in healthcare ranging  from biomedical text mining to medical question answering \cite{sharaf2023analysislargelanguagemodels, 21976}. Some larger domain related models like BioBERT \cite{Lee_2019} where fine-tuned on multiple medical related tasks to serve in such applications \cite{sharaf2023analysislargelanguagemodels}. Other approaches used generic LLMs with Zero Shot approaches \cite{Plagwitz2024, tang2024medagentslargelanguagemodels}. Nowadays, it is less common for an LLM to be fine-tuned to a specific task. While fine-tuning LMMs can improve their performance on the training-specific domain, the process of fine-tuning a model on specific datasets is nether standardized nor straightforward. Though there are some projects that attempt to address this issue \cite{axolotl, unsloth} by streamlining the training.
In the medical context two additional challenges arise when attempting to fine-tune a model: The potential training data can be limited in amount. This can impair fine-tuning as it tends to yield larger improvements when provided with large amounts of training data. Another complicating aspect can be the sensitive nature of medical data. Some datasets can not be given to external services for processing or training purposes, making it necessary for researchers to create a local setup for the processing of data and the training of models.
In this work we looked into these two limitations and asked: How beneficial is locally hosted fine-tuning on small medical datasets to natural language processing tasks? Specifically, we look into text classification and named entity recognition (NER).
For text classification we use the data from the i2b2 Smoking Challenge \cite{UZUNER200814} as an example for medical texts and evaluate the performance of various models and compared their performance to models fine-tuned on progressively larger amounts of data. In a similar way we evaluate the performance of fine-tuned models on the NER task proposed in \cite{Plagwitz2024} where entities of interest were extracted from German cardiology reports.

\section{Materials and Methods}
\label{sec:materials}
\subsection{Data set - i2b2 Smoking Challenge (English)}
As a classification task we choose the “smoking challenge” proposed in \cite{UZUNER200814} for this study. The challenge was to classify the smoking status of patients based on their de-identified discharge summaries. All summaries were annotated by pulmonologists with one of five categories: Past Smoker, Current Smoker, Smoker, Non-Smoker, Unknown. In the challenge the data was split up in 398 training- summaries and 104 test-summaries. We split the training-set into four (nearly) equally sized folds, maintaining the global distribution of ground truth labels. From these folds, we considered five different training sizes, consisting of half a fold, one fold, two folds, three folds, and all four folds. These sets were then used to analyze the model’s performance as a function of the training data size.
\subsection{Data set – Cardiology Reports (German)}
As a NER task we used the setup described in \cite{Plagwitz2024}. That work attempted NER on German medical text, aiming to extract values for specific medical entities from a set of cardiology MRI reports. The training set consisted of 498 annotated samples, which we randomly split in five folds. From four of the folds, we created four sets of training data as described for the smoking challenge data, leaving out the half fold. The remaining fold served as the evaluation-set. 
\subsection{System infrastructure}
We trained the llama3-8b-instruct model \cite{llama3modelcard} using the Axolotl \cite{axolotl} framework. The model was trained using Low-Rank Adaptation (LoRA) for one epoch with each of the different sized training-datasets. We trained for only one epoch to show that results can be improved with minimal effort and to avoid potential problems with overfitting. The trained adapters were merged with the model and converted to the GGUF file format using llama.cpp \cite{llamacpp}. The merged models were uploaded to an Ollama \cite{ollama} instance, which we used for inference.
\subsection{Query procedure}
Each query send to a model contained a prompt consisting of three parts: 'system', 'user', and 'output'. For the classification task, the system prompt defined the five possible (smoking status) classes and instructed the model to respond in Python-readable JSON format. In contrast, the system prompt for the NER task followed the approach outlined in \cite{Plagwitz2024}, specifying the entities to extract and defining a default value for non-occurring entities. Both tasks used a similar structure, with the user prompt providing context and input text, and the output containing the predicted class or extracted entities in JSON format. We evaluated each dataset by making API calls to an Ollama instance with zero temperature and enforcing JSON responses.
\subsection{Training procedure}
We created Axolotl configuration files for each fine-tuning setup, more precise five for the smoking challenge and four for the cardiology reports. As the basis we used Meta’s llama3-8b-instruct model, loading it in 8 bit for memory and training efficiency.

\section{Evaluation}
For the classification task we used three evaluation metrics: parsing error rate, classification accuracy, and F1 score. A parsing error occurred when the model response couldn't be read or lacked the "status" key after extracting the Python dictionary (identified as the first substring matching "{...}"). Correct classifications required matching values between the predicted and the annotated status; otherwise, it was classified as an additional class 'fails'. This allowed us to compute accuracy and F1 scores.
We evaluated the performance on NER tasks using four metrics: accuracy per text, accuracy per entity, F1 score, and parsing error rate. To assess parsing errors, we compared the model's output dictionaries against the expected entity names (see \cite{Plagwitz2024}), their presence as the dictionary keys indicating success. Differing dictionary keys indicated a parsing error. Responses with parsing errors were subsequently considered incorrect. The accuracy was measured on to levels: accuracy per text measured the proportion of entirely correct predictions, whereas accuracy per entity gauged the fraction of correctly identified values across all instances of each entity. Finally, we computed a weighted F1 score by assigning binary labels ("true" or "false") to match individual entity values against their corresponding gold standard annotations.
\section{Results}
On the classification task our trained model performed better on accuracy and f1 score the more training data they had available, as shown in Table \ref{tab:model-performance-smoking}. In our experiment 200 training data points were enough to outperform, the zero shot use of existing models, even the one that was multiple times larger than our training models. 
On the named entity recognition task our fine-tuned models reached the performance of bigger existing open-source models as shown in Table \ref{tab:model-performance-cardiology}. Our trained model performances improved again with more training data, seemingly plateauing around 300 training data points. The model trained on 300 training data points performed on par with the biggest open-source model evaluated. In contrast to the classification task, we saw an improvement regarding the parsing error.

\begin{table}
\caption{Performance of different open-source models on the 104 English medical records reserved for evaluation from the i2b2 smoking challenge data. “Dummy” indicates the performance of a classifier that classifies all texts as “UNKNOWN”.}
\centering
\begin{tabular}{|l|l|l|l|l|l|l|l|l|}
\hline
Model names                & Params & Training data & Accuracy of classification & F1 score & Parsing errors \\
\hline
Dummy                      & *      & *             & 0.606                    & 0.457    & *            \\
Llama3 instruct            & 8b     & *             & 0.327                    & 0.330    & 0            \\
Llama3 instruct trained    & 8b     & 53            & 0.413                    & 0.391    & 0            \\
Llama3 instruct trained    & 8b     & 101           & 0.615                    & 0.619    & 0            \\
Llama3 instruct trained    & 8b     & 201           & 0.856                    & 0.851    & 0            \\
Llama3 instruct trained    & 8b     & 300           & 0.885                    & 0.883    & 2            \\
Llama3 instruct trained    & 8b     & 398           & \textbf{0.913}           & \textbf{0.906} & 0            \\
Gemma instruct             & 9b     & *             & 0.740                    & 0.747    & 1            \\
Gemma2                     & 9b     & *             & 0.76                     & 0.744    & 2            \\
Llama3 instruct            & 70b    & *             & 0.625                    & 0.65     & 0            \\
\hline
\end{tabular}
\label{tab:model-performance-smoking}
\end{table}

\begin{table}
\caption{Performance of different open-source models on the 100 German cardiology reports reserved for evaluation. Many of the reports do not contain a correlating value to one or more of the entities we aim to extract. These not findable entity values where annotated as “-”. The “Dummy” indicates the performance of an extractor that returns “-” as the value for each entity on all texts.}
\centering
\begin{tabular}{|l|l|l|l|l|l|l|l|l|}
\hline
Model names                & Params & Training data & Accuracy per text & Accuracy per entity & F1 score & Parsing errors \\
\hline
Dummy                      & *      & *             & 0.02              & 0.558               & 0.637    & *            \\
Llama3 instruct            & 8b     & *             & 0.77              & 0.825               & 0.904    & 17           \\
Llama3 instruct trained    & 8b     & 100           & 0.77              & 0.921               & 0.958    & 2            \\
Llama3 instruct trained    & 8b     & 200           & 0.8               & 0.859               & 0.924    & 13           \\
Llama3 instruct trained    & 8b     & 300           & 0.94              & \textbf{0.986}      & \textbf{0.993} & 1            \\
Llama3 instruct trained    & 8b     & 398           & 0.9               & \textbf{0.987}      & \textbf{0.993} & 0            \\
Gemma instruct             & 9b     & *             & 0.9               & 0.983               & 0.991    & 0            \\
Gemma2                     & 9b     & *             & 0.67              & 0.959               & 0.978    & 0            \\
Llama3 instruct            & 70b    & *             & \textbf{0.95}     & \textbf{0.986}      & \textbf{0.993} & 1            \\
\hline
\end{tabular}
\label{tab:model-performance-cardiology}
\end{table}

\section{Discussion}
Our research demonstrates the potential of fine-tuned models on tasks like text classification and NER. Other recent research regarding LLMs in medicine were often directed towards creating models trained for medical question answering \cite{oneLLMisNotENough,enhancingLLMs}. In contrast to those more general approaches, we use LLMs as tools for highly specific tasks. We show empirically that training LLMs locally on small sensitive datasets can yield comparative results. Our fine-tuned models were able to nearly match larger models on the NER task with training sets of 300 data points and surpass them on the classification task using training sets of 200 data points. Training also increased the adherence to a machine-readable output format. These results were produced on English and German texts, indicating that fine-tuning could be useful in compensating for potential language biases present in LLMs mostly trained on English texts. Based on these observations we believe that fine-tuning LLMs for task-specific applications holds great promise for extracting structured data from unstructured text. Applications using task specific LLMs could save time and effort of personal working in healthcare.

\section{Conclusion}
Our case study demonstrates the effectiveness of fine-tuning LLMs on small medical datasets for text classification and NER tasks. Despite the limited availability of training data, our locally fine-tuned models achieved competitive results with larger models, including models with notably more parameters. Our findings suggest that fine-tuning LLMs on as little as 200-300 training examples can lead to substantial improvements in performance, particularly for text classification tasks. Overall, our study provides evidence that task-specific fine-tuning of LLMs has significant potential for improving the automation of clinical workflows and facilitating the efficient extraction of structured data from unstructured medical text.

\section*{Declarations}
The authors declare, that there is no conflict of interest. 
All authors approved the manuscript in the submitted version and take responsibility for the scientific integrity of the work.
Author contributions: NL, LP: conception of the work; LP,AB data acquisition and interpretation; NL: data analysis and interpretation; NL: writing the manuscript, LP, AB, JV substantial revising of the manuscript. All authors approved the manuscript in the submitted version and take responsibility for the scientific integrity of the work.
Acknowledgement: We would like to thank the authors of the 2006 n2c2 datasets "Deidentification \& Smoking" for sharing this valuable research resource that has made this study possible.
Funding: This work was supported by the Interdisciplinary Center for Clinical Research (IZKF) Münster, Germany (grant SEED/020/23 to AB).

\bibliographystyle{unsrt}  
\bibliography{references}  

\end{document}